\documentclass[sigconf]{acmart}
\usepackage{graphicx}
\usepackage{multirow}

\AtBeginDocument{%
  \providecommand\BibTeX{{%
    \normalfont B\kern-0.5em{\scshape i\kern-0.25em b}\kern-0.8em\TeX}}}

\copyrightyear{2022}
\acmYear{2022}
\setcopyright{acmlicensed}
\acmConference[KDD 22]{KDD '22: ACM SIGKDD International Conference on Knowledge Discovery \& Data Mining}{August 14--18, 2022}{Washington, D.C. }
\acmPrice{15.00}
\acmDOI{10.1145/1122445.1122456}
\acmISBN{978-1-4503-9999-9/18/06}
\acmPrice{15.00}
\acmISBN{978-1-4503-XXXX-X/18/06}

\begin{document}

\title{Multiple Attribute Fairness: Application to Fraud Detection}


\author{Meghanath M Y}
\affiliation{%
  \institution{Adobe}
  \city{San Jose}
  \country{USA}
}
\email{yadagiri@adobe.com}
\author{Sriram Ravindran}
\affiliation{%
  \institution{Adobe}
  \city{San Jose}
  \country{USA}
}
\email{sravindr@adobe.com}

\author{Deepak Pai}
\affiliation{%
  \institution{Adobe}
  \city{San Jose}
  \country{USA}
}
\email{dpai@adobe.com}

\author{Anish Narang}
\affiliation{%
  \institution{Adobe}
  \city{San Jose}
  \country{USA}
}
\email{annarang@adobe.com}

\author{Vijay Srivastava}
\affiliation{%
  \institution{Adobe}
  \city{San Jose}
  \country{USA}
}
\email{vijays@adobe.com}
%

\renewcommand{\shortauthors}{Meghanath et. al.}

\begin{abstract}
We propose a fairness measure relaxing the equality conditions in the popular equal odds fairness regime \cite{hardt2016equality} for classification. We design an iterative, model-agnostic, grid-based heuristic that calibrates the outcomes per sensitive attribute value to conform to the measure. 
 The heuristic is designed to handle high arity attribute values and performs a per attribute sanitization of outcomes across different protected attribute values. We also extend our heuristic for multiple attributes. Highlighting our motivating application, fraud detection, we show that the proposed heuristic is able to achieve fairness across multiple values of a single protected attribute, multiple protected attributes. When compared to current fairness techniques, that focus on two groups, we achieve comparable performance across several public data sets. 

\end{abstract}




\begin{teaserfigure}
\vspace{-1.5cm}
\centering
\includegraphics[width=0.9\textwidth]{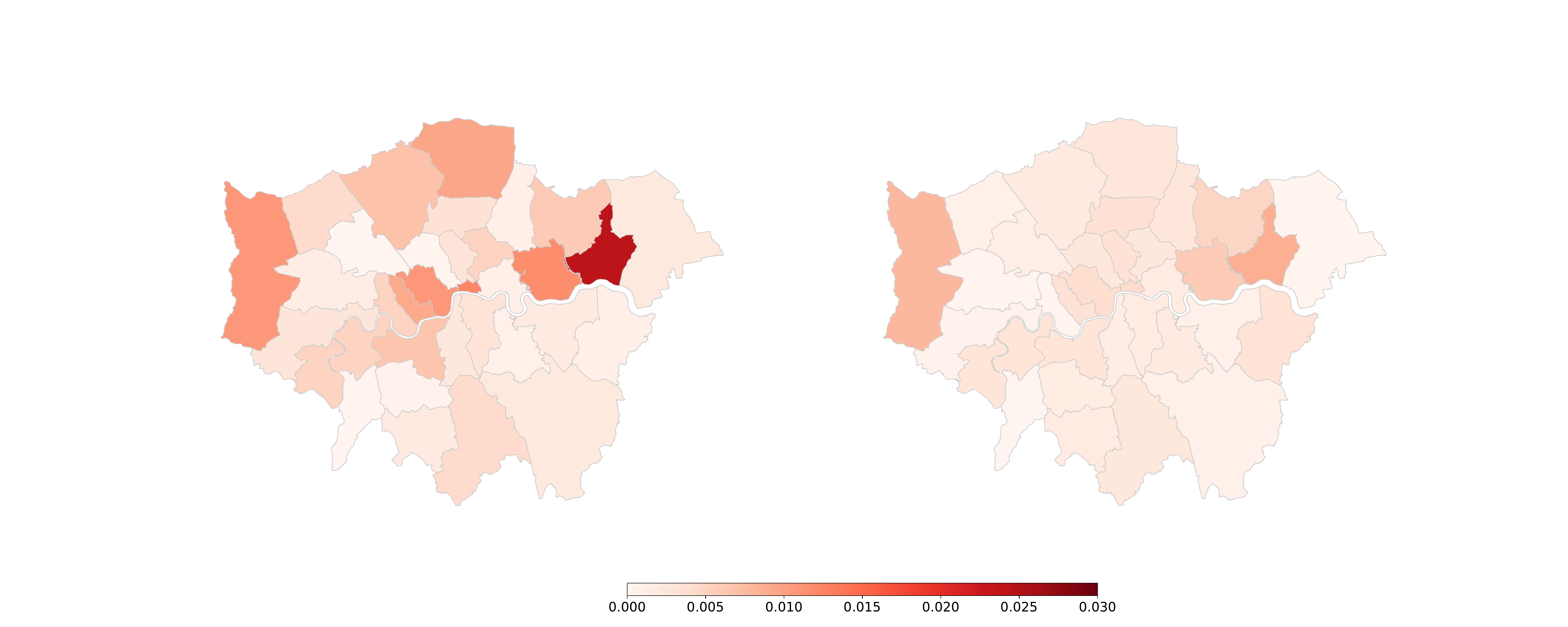}
\caption{Fraud Detection : False Positive Rates Pre (Left) and Post (Right) Sanitization.}
\label{fig:countryprepost}
\end{teaserfigure}
\maketitle

\begin{figure*}[t]

\end{figure*}
\vspace{-0.1in}
\section{Introduction}
Machine learning algorithms are increasingly being used to make decisions in critical areas such as access to credit, employment opportunities and education. Recent research has highlighted concerns related to potential bias, where ML techniques produce discriminatory outcomes \cite{chouldechova2017fair}. For instance, when Apple released its credit card, there were claims that women were given a lower credit limit. Similarly, ProPublica analyzed a risk assessment used by judges in the United States to predict recidivism risk for an accused and concluded that the predictions are biased against black defendants . From an industry standpoint, it is critical to sanitize ML algorithms, ensure fairness and avoid ethical and legal concerns. 

Detecting and mitigating bias in ML algorithms has gained traction in the last few years both in academia and industry. However, existing approaches have certain limitations, primarily around accounting for ground truth distributions as well as type-1 and type-2 errors. In this paper, we use online transaction fraud detection as a motivating example to discuss the drawbacks of the current approach and to illustrate the value of the proposed approach.

\textit{Motivating Application} : As businesses continue to shift toward online payments, there is a rising need to have an effective fraud detection solution capable of real-time, actionable alerts and insights. This increasing need has led to several solutions being offered by industry leaders \cite{gartner} such as Amazon \cite{amazonfrauddetector}, Microsoft \cite{microsoftfrauddetector}, ThreatMetrix \cite{threatmatrix}, Forter \cite{forter}. These solutions can analyze hundreds of parameters of an online transaction such as transaction amount, past transaction trends, IP location of the transaction, transaction time, merchant name etc. and output a score in real-time (usually from 0 to 100) indicating the fraudulent nature of the transaction. While these solutions claim a high predictive performance, biases in the training data and model inaccuracies can lead to decisions that treat individuals unfavorably (unfairly) based on characteristics such as transaction country and currency (sensitive attributes). Ensuring that these biases are curtailed before deploying such solutions is critical for the business. 

In this work, we propose a fairness measure for classification and detail a heuristic based on model calibration to ensure such fairness in the outcomes. We showcase the efficacy of the proposed approach on several publicly available data sets and a proprietary commercial data set that is comprised of online transactions (Figure \ref{fig:countryprepost}). For baselines, we consider Equalized Odds \cite{hardt2016equality} and Calibrated Equalized Odds \cite{pleiss2017fairness}.

\vspace{-0.1in}
\section{Related Work}

\begin{figure}[!h]
\vspace{-0.1in}
\centering
\includegraphics[width = 0.8\linewidth]{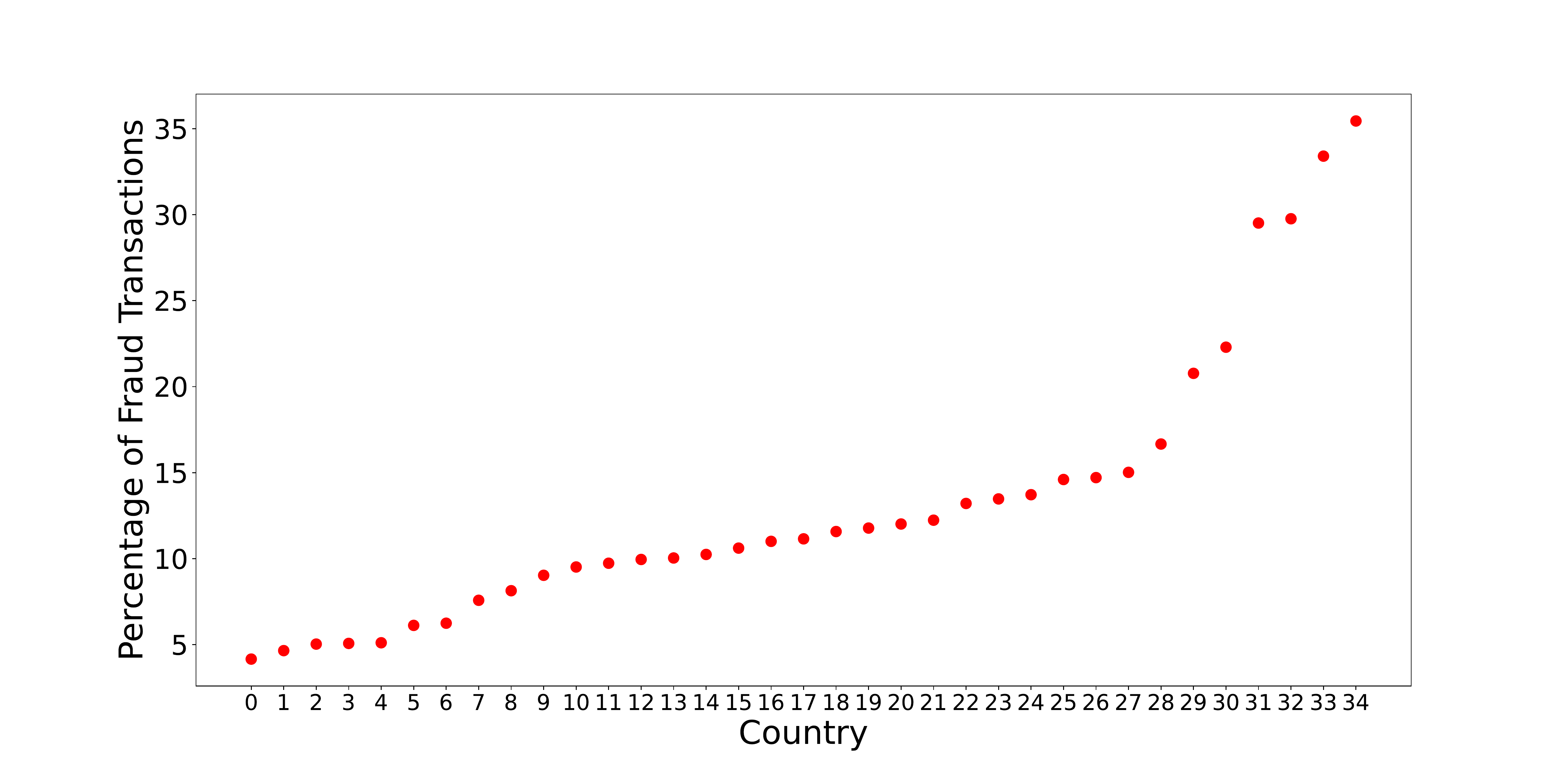}
\caption{Ground Truth Fraud Rates across Countries.}
\label{fig:gtruthcountry}
\vspace{-0.1in}
\end{figure}

Anti-discrimination laws in the United States regulate two discriminatory behaviors based on protected attributes. These doctrines of discrimination are disparate treatment and disparate impact. Disparate treatment protects individuals who are affected by algorithmic decision making against explicit discrimination. The standard practice to conform to disparate treatment in Machine learning techniques used for decision making is to exclude protected attributes from inputs. This is easy to achieve in a fraud detection model – one would just need to not use any sensitive attribute while training the model.
In contrast, disparate impact addresses outcome discrimination; it recognizes liability for practices with uneven impacts on different protected groups. While there is no single fairness measure that captures absolute impact parity, equal opportunity, demographic parity, and equalized odds \cite{beutel2017data, hardt2016equality, chzhen2019leveraging, pleiss2017fairness} have received considerable attention.
Efforts from scholars revolve around heuristics to ensure these fairness measures. While these techniques are catered to ensure fairness for any Machine learning technique, they are not always practical to implement in an industry setting. 
To expound, consider demographic parity, which requires that the outcome rates across sensitive attributes be the same. In the context of fraud detection, for a sensitive attribute “country”, this would mean that acceptance or rejection rates of online transactions for each country should be the same. This may not be ideal since we observe that the ground truth fraud rates from different countries are significantly different (Refer to Figure \ref{fig:gtruthcountry}). This could possibly be explained by the jurisdiction in each of these countries in handling online fraud (stricter penalties would mean lower fraud rates and vice versa). If we were to sanitize our fraud detection model to ensure demographic parity, one would either end up allowing lots of fraudsters or declining lots of genuine users both of which are not an ideal outcome for a firm.

Another approach, equalized odds, aims to achieve the same false positive and true positive rates across sensitive attributes. While this is better than the demographic parity for fraud detection, the measure enforces equal importance to both false positives and false negatives. This is again not ideal for a firm given the cost of type-1 and type-2 errors are different for electronic\footnote{The cost of declining a genuine customer is much higher than allowing a fraudster.} vs physical goods. Lastly, Equalized opportunity, is a relaxed version of Equalized odds that only requires equal true positive rates across sensitive attributes. This again is insufficient since we would not want to deny a genuine customer and hence would like to balance false positive rates as well. Furthermore, while getting to \textit{equal} rates is possible in theory, it is less practical with daily fluctuations in traffic and fraud patterns.
Additionally, from an implementation standpoint, another major critique about the current techniques is that they require multiple sanitizations of the ML model to achieve fairness. This largely arises since majority of these techniques assume binary sensitive attributes. Such constraints are not viable in a fraud detection system that usually deal with high arity attributes and require real time inference. Further, to the best of our knowledge, none of these methods perform sanitization of a machine learning model in a post-hoc fashion across multiple protected attributes. 
We tackle these drawbacks in our work by introducing a relaxed Equalized odds fairness measure and a one-shot fairness heuristic to achieve the proposed fairness measure. Further, we design the proposed heuristic to have the flexibility to pick either or both constraints of the Equalized Odds definition (similar FPR or similar TPR or both) \cite{hardt2016equality}. Finally, and importantly, we also extend the proposed fairness measure and heuristic to handle multiple attributes.

\vspace{-0.1in}
\section{Method}
The classical definition of Equality of Odds posits that true positive  rates and false positive rates across a protected attribute are the same. It reflects a fundamental idea of fairness, that qualified individuals should be given equal opportunity (true positive rate - TPR) to access a desirable outcome while also requiring equal false positive rates (FPR, ensuring no bias to the positive class) regardless of their demographics or any other sensitive attributes. However, this strict requirement of equality in most settings requires non-deterministic thresholds as pointed out in previous research \cite{beutel2017data, hardt2016equality, chzhen2019leveraging, pleiss2017fairness} making it less appealing from a practical standpoint. Further, the current heuristics to achieve such fairness involves multiple sanitizations – one per each categorical attribute value (Ex: one for each country) of a machine learning model to ensure the fairness criterion is met. 
We deal with these issues by first, relaxing the notion of the standard Equalized odds. Specifically, we relax the equality constraint of FPR, TPR and enforce a weaker, yet practical constraint. Second, we propose a fairness heuristic that sanitizes the outputs of a classification model to conform to the relaxed equalized odds measure. \\
\textbf{Note}: We describe the details assuming a single attribute and later extend this to multiple attributes in Section \ref{sec:multiple}. 
\vspace{-0.1in}
\subsection{Relaxed Equalized Odds}
Say $D=\{d_1,d_2,…,d_K\}$ is a protected attribute with arity $K$, we say that a fraud detector model $F$\footnote{In general, $F$ can also be any classification model that needs to be sanitized for bias.} satisfies relaxed equalized odds with respect to attribute $D$ if the false positive and true positive rates of the respective attribute values $fpr^F=\{fpr_1^F,fpr_2^F,…,fpr_K^F\}$, $tpr^F=\{tpr_1^F,tpr_2^F,…,tpr_K^F\}$ lie within $n$ standard deviations of their means i.e.,  
\vspace{-0.1in}
\begin{align}
    \mu(fpr^F )- n \sigma(fpr^F )\leq fpr_i^F \leq \mu(fpr^F )+  n \sigma(fpr^F) \\ 
    \mu(tpr^F )- n \sigma(tpr^F ) \leq tpr_i^F \leq \mu(tpr^F )+ n \sigma(tpr^F ) 
\end{align}
\vspace{-0.2in}

where $\mu$ and $\sigma$ are the average and standard deviation respectively across the $K$ attribute values. 

The above definition while weaker, captures the core philosophy of the Equalized odds in a high arity($K$)\footnote{Note that mean and standard deviation across the attributes would only hold a statistical meaning if the number of attribute values are higher. This is traditionally the case in fraud detection models where a sensitive attribute, say country or currency could take over 100 types of values. }  setting – similar opportunity and similar false alarms across a sensitive attribute.
\vspace{-0.1in}
\subsection{Fairness Heuristic}
Following a similar notation as above, say we would like to sanitize a fraud detector $F$ with respect to the protected attribute $D$, our fairness heuristic has the following key steps. 
    \subsubsection{Choice of constraints} We provide an end user the capability to further relax the traditional equalized odds definition to conform to either similar FPRs or TPRs or both (i.e., Eq 1 or Eq 2 or both). This would be useful for a business where false positives of a fraud detector are more costly compared to true positives. Said differently, if it is costlier for a business to accept a fraudulent customer compared to declining a genuine customer (false negative), then they might choose to conform to similar FPRs, while also ensuring that the overall FPR is low. In a similar way, a business might choose to have equal opportunity (TPR) across a protected attribute. 
    This choice decides the $selection_{metric}$ in our heuristic. $selection_{metric}  = F1$; if Eq (1) and Eq (2), $= F0.5 ;$, if only Eq (1) and $= F2 ;$ if only Eq (2). The choice of the selection metric is done based on the importance of the FPR and TPR to the end user. F1 treats false positives and false negatives (true positives) equally while $F0.5$ and $F2$ weight false positives and false negatives higher respectively.
	\subsubsection{Threshold grid initialization} The core idea of our heuristic is to calibrate the decision thresholds of the model $F$ across the different attribute values D to conform to Eq (1) and/or Eq (2). To do this, we initialize a linear grid of possible threshold values denoted as $G_{thresh}$. A choice of $G_{thresh}$  could be $\{0.6,0.61,…,0.9\}$. Note that depending on the threshold value, both FPR and TPR of the model $F$ would change.
	\subsubsection{Performance computation} For all the values in $G_{thresh}$, per attribute value in $D$, we compute model $F$’s performance metrics. These comprise of the metric that the end user chooses to conform to in the fairness measure (say FPR in our running example). At the end of this step, we would have $fpr^{F_g}=\{fpr_1^{F_g},fpr_2^{F_g},…,fpr_K^{F_g}\}$  $\forall$ $g \in G_{thresh}$, where $fpr^{F_g}$ is the FPRs of the model $F$ at threshold $g$ across protected attribute values. 
	\subsubsection{Pruning and Selection} Our next task is to iteratively prune the set of $\{fpr^{F_g},g \in G_{thresh} \}$ to conform to Eq 1. and select custom thresholds from $G_{thresh}$ for each attribute in $D$.  \\
\textbf{Prune} : To do this, we compute $\mu(fpr^{F_g})$  and $\sigma(fpr^{F_g})$, $\forall$  $g \in G_{thresh}$ and prune $fpr_i^{F_g}$ that are overly biased compared to the mean, i.e., violate the constraint $fpr_i^{F_g} \leq \mu(fpr^{F_g}) + n \sigma(fpr^{F_g})$. This would mean that for a protected attribute value $d_i$, $D=\{d_1,d_2,…,d_K\}$, some of the threshold choices are pruned since their false positive rates   are statistically higher than the average. Denote such pruned set as $\{fpr^{F_g}\}^{pruned}$.\\ 
	    \textbf{Select} : From $\{fpr^{F_g}\}^{pruned}$, we next select the threshold per each attribute value $d_i$ where the $selection_{metric}$ (F-0.5 in our running example) is maximized. This results in a choice of $\{(d_1,g_1 ),...,(d_K,g_K)\}$. Denote this choice of thresholds as the model $F_{select}$ \\
	   We check if $F_{select}$ conforms to the fairness constraints, i.e., Eq (1) in our running example
$\mu(fpr^{F_{select}}) - n\sigma(fpr^{F_{select}} ) \leq fpr_i^{F_{select}} \leq \mu(fpr^{F_{select}})+ n\sigma({F_{select}})$. If not, we repeat the pruning and selection step with the updated $\{fpr^{F_g}\}^{pruned}$ until the constraint is satisfied. 
First, we note that the final $F_{select}$ conforms to the fairness measure and preserves model fidelity by selecting the best available model based on the selection metric. Second, from Step 3 and Step 4, we note that our fairness heuristic only requires aggregate information such as FPR, TPR, F-1 score about the protected attribute $D$ and does not require information about the features (protected or unprotected) or their feature values used to train the fraud detector model. Hence, one could perform this sanitization in a differentially private manner by adding appropriate noise to the aggregate information. Third, the reliance of the output on solely the outcomes makes the fairness heuristic model agnostic. Fourth, the proposed heuristic is a one-shot search heuristic that conforms the model $F$ to the protected attribute $D$ rather than performing $K$ sanitizations for each attribute of $D$. Finally, we note that the grid heuristic discussed above is guaranteed to converge to a solution if there is indeed a solution where the FPRs/TPRs are equal for the sub-groups.
\vspace{-0.2in}
\subsection{Multiple Attribute Extension}\label{sec:multiple}
The fairness heuristic detailed above handles a single protected attribute and conforms to equalized fairness in that attribute. However, for most practical applications, we would require a machine learning model to conform to fairness across multiple attributes. For instance, in the case of a fraud detection model, one would like to sanitize the model to not be biased against a country, currency that they use. The current state of the art fairness heuristics deals with this issue while training \cite{foulds2020intersectional} or by processing the features before learning \cite{zhang2021omnifair}. Supplementing this line of literature, we extend the proposed fairness heuristic which performs sanitization post inference to handle multiple attributes. 

A naive way to handle multiple attributes (consider two for example) say $D^1=\{d_1^1,d_2^1,…,d_{K^1}^1 \}$, $D^2=\{d_1^2,d_2^2,…,d_{K^2}^2\}$ in our regime would be to identify $\{((d_1^1  d_1^2),g_1), ... ,(d_{K^1}^1  d_{K^2}^2), g_{K^1 \times K^2})\}$. Note that such calibration of thresholds would conform to a strict notion of the Relaxed Equalized Odds ensuring the FPR and TPRs are equal across every sub-population. While this is certainly an attractive property to have, the computational complexity is exponential. A weaker notion would be to ensure that the FPR and TPR values satisfy the constraints in Eq (1) and (2) per attribute independently rather than all enumerations. 

Expounding the two concepts with an example of two attributes country and currency with attribute values US, IN and USD, INR respectively. In the stronger notion, we want FPRs and TPRs across the eight sub populations $\{(US,USD),(US,INR),(IN,USD),\\ (IN,INR),(US),(USD),(IN),(INR)\}$ to conform to Eq (1) and (2). In the weaker notion, we want them to conform to four sub populations $\{(US),(IN),(USD),(INR)\}$. While the former is a stronger fairness notion and scales exponentially with the number of attributes, the latter is weaker and scales linearly.
We propose heuristics for both such notions of fairness giving the decision maker the flexibility to opt to either of those depending on the number of attributes the model needs to be sanitized on. We recommend the stronger notion for 10 or a smaller number of attributes and the univariate, weaker notion for greater than that.  
In both cases, we reduce this combinatorial computation by first pruning the attribute space by identifying attributes that encompass similar sub populations. 

\subsubsection{Attribute Pruning} 
If two attributes are dependent i.e., capture similar sub populations with their attribute values (for instance country and currency), it is redundant to calibrate the thresholds for both. We define dependence between attributes as the Chi-square statistic between pairs of attributes. Based on the statistic, we compute the $p$-value to infer if two attributes are statistically independent of each other (p-value <= 0.01) and drop the attributes that are highly dependent. This step prunes the space of protected attributes while only leaving independent attributes to calibrate.

\subsubsection{Strong Multiple Attribute Fairness : Attribute Value Pruning}
In this regime, we want to ensure that the FPRs and TPRs of every sub-population conform to Eq (1) and (2). While reducing the attribute space partly addresses the computation issue discussed earlier, we are still left with potentially multiple attributes with high arity. For instance, say we have m protected attributes after pruning with cardinalities $K_1,K_2,…,K_m$, we would be looking at possibly $K_1×K_2×…×K_m$ subspaces to calibrate the thresholds on. However, we note that to yield a reliable estimate of the performance metrics fpr,tpr  (Step 3 in the heuristic) or the selection metrics (Step 4.b.), as a rule of thumb \cite{green1991many}, a subspace would at least need to have 100 transactions (data points). Hence, we prune away such subspaces using a scalable implementation of the Frequent Pattern Tree \cite{han2000mining} data structure. 
Let $C=\{c_1,c_2,…,c_P\}$ be the total number of subspaces left with each at least having 100 transactions. An example of a $c_1$ would be $\{country=US,sex=Male,race=Hispanic\}$. The resulting clusters $C$ are then calibrated with the grid of thresholds to conform to the Relaxed Equalized Odds fairness measure in a similar fashion to the single attribute case where the performance table looks like $\{fpr^{F_g}\}, g \in G$. 
\subsubsection{Weak Multiple Attribute Fairness }
In this regime, we want to ensure that the FPRs and TPRs per attribute conform to Eq (1) and Eq (2). Operationally, this can be achieved by changing the Step 3) in the fairness heuristic to compute the performance table across all attributes and their attribute values. That is, for m protected attributes, construct
$fpr^{F_g}=\{fpr_1^{F_g},fpr_2^{F_g},…,fpr_{K_1}^{F_g}, fpr_1^{F_g}, fpr_2^{F_g},…,fpr_{K_2}^{F_g},\ldots \\ fpr_{K_{1+K_2+ \ldots K_m}}^{F_g}\}$  $g\in G_{thresh}$, where $fpr^{F_g}$ is the FPRs of the model $F$ at threshold g across all univariate enumerations of the protected attribute values. This is followed with the Pruning and Selection step as discussed earlier to find thresholds per attribute value for all the m attributes. 

\vspace{-0.1in}
\section{Experiments}

\begin{figure}[!h]
\vspace{-0.2in}
\centering
\includegraphics[width = 0.8\linewidth]{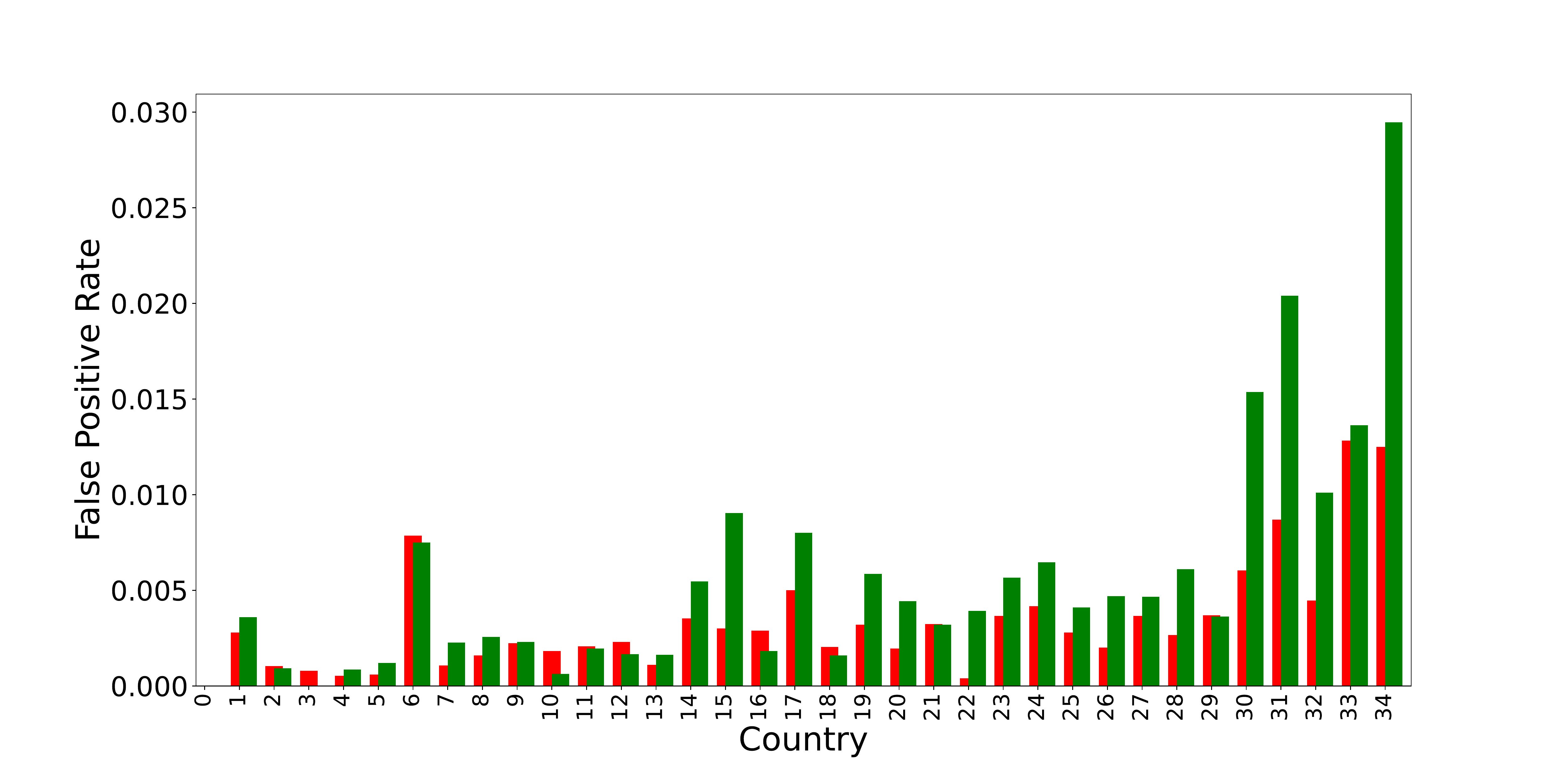}
\vspace{-0.2in}
\caption{Fraud Detection FPR (Pre = Red, Post = Green)}
\label{fig:sanfprbar}
\vspace{-0.2in}
\end{figure}
We showcase the efficacy of the proposed approach  on several publicly available data sets - criminal recidivism\footnote{Data source : https://github.com/propublica/compas-analysis}, income-prediction , health prediction\footnote{Data source : https://github.com/gpleiss/equalized\_odds\_and\_calibration/}, and a proprietary commercial data set that is comprised of online transactions. For baselines, we consider Equalized Odds \cite{hardt2016equality} and Calibrated Equalized Odds \cite{pleiss2017fairness}. In the experiments, we set $n$ = 2 and investigate single protected attribute with high arity, multiple protected attributes, and also consider the case of 2 protected groups as investigated by earlier studies. 

\begin{figure}[!h]
\vspace{-0.2in}
\centering
\includegraphics[width = 0.8\linewidth]{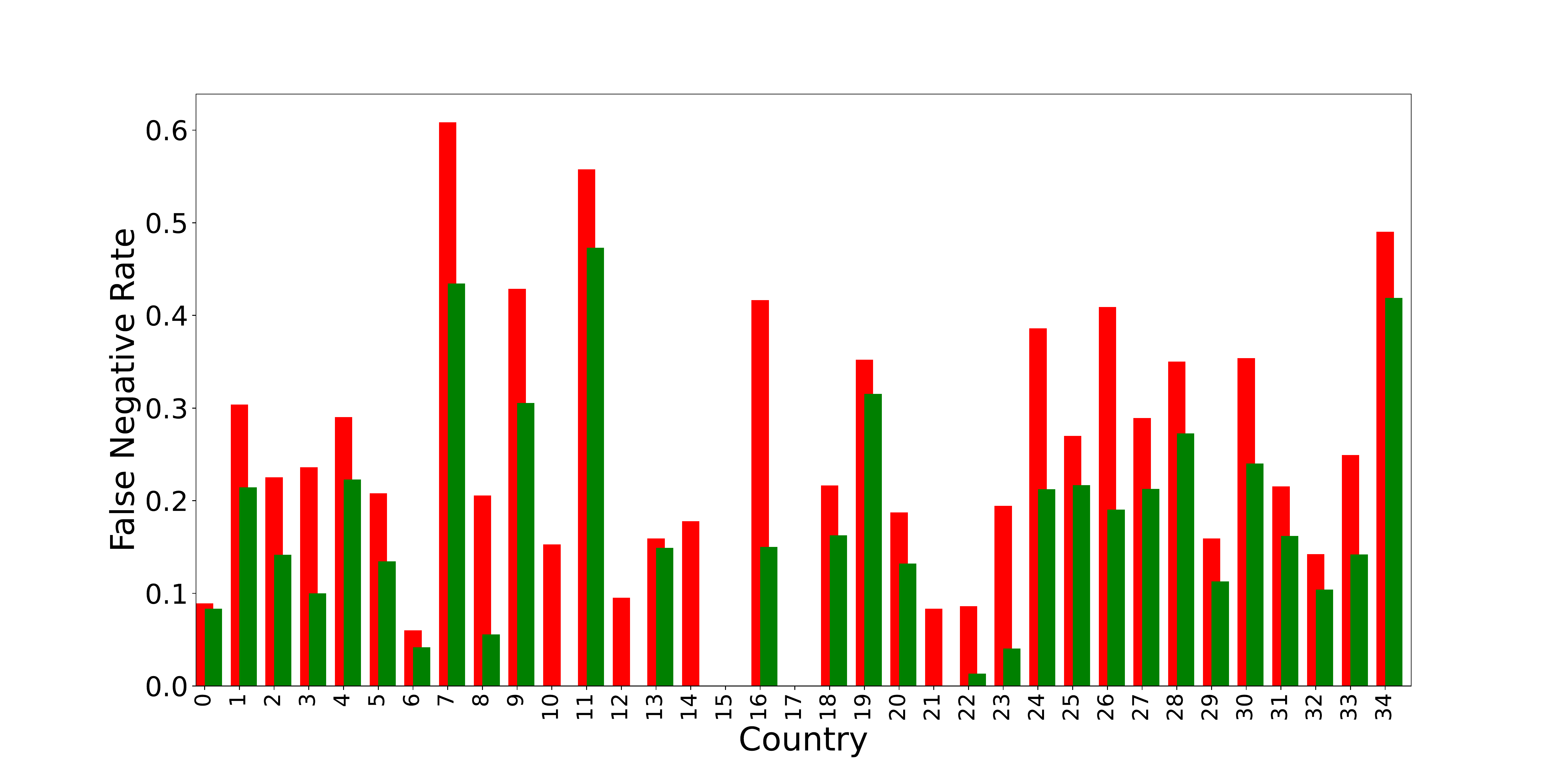}
\vspace{-0.2in}
\caption{Fraud Detection FNR (Pre = Red, Post = Green)}
\label{fig:sanfnrbar}
\vspace{-0.2in}
\end{figure}

\textbf{Case Study : Fraud Detection} We first discuss our motivating example, online fraud detection\footnote{In all the experiments, positive refers to a fraudulent transaction. Hence true positives would mean fraudulent transaction being identified as fraud by the detector.}. The aim of the prediction is to assess whether an online transaction initiated by an individual is fraudulent. The prediction model omits all protected attributes as input to conform to the disparate treatment doctrine of fairness measure. However, as evident from \ref{fig:sanfprbar} and \ref{fig:sanfnrbar} (red color bars) we observe that model exhibits a bias towards the demographic attribute country. A similar trend is observed when we look at the FNRs. As we motivate earlier, from a business perspective, it is useful to have flexibility to sanitize these metrics across attributes

We employ the proposed heuristic on the predictions made by the fraud detector by conforming to both the constraints Eq (1) and Eq (2). From Figures \ref{fig:sanfprbar}, \ref{fig:sanfnrbar} (green color bars), we observe that the final $F_{select}$ with custom thresholds per country considerably reduces the bias towards certain countries. Further, we also note that the FPRs and FNRs across countries conform to the relaxed equalized odds fairness measure -- mean across the countries lies withing two standard deviations.



\begin{figure}[!h]
\vspace{-0.2in}
\centering
\includegraphics[width = 0.8\linewidth]{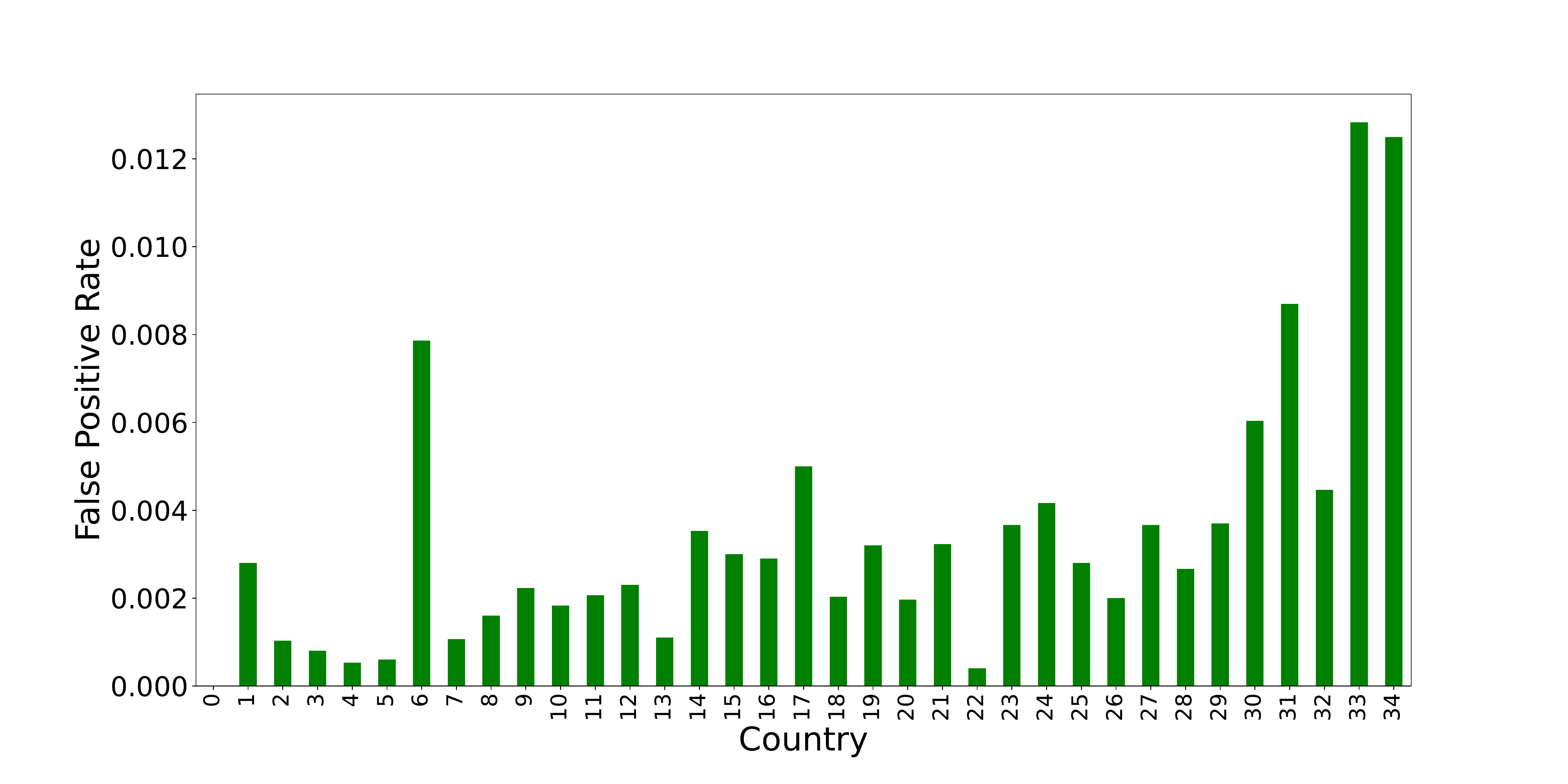}
\vspace{-0.2in}
\caption{Fraud Det. (XGBoost, Multiple attributes) FPR (Post)}
\label{fig:sanxgboostfprbar}
\vspace{-0.2in}
\end{figure}

\begin{table*}[]
\resizebox{0.8\linewidth}{!}{%
\begin{tabular}{|c|clll|clll|clll|}
\hline
\centering
\multirow{2}{*}{\textbf{\begin{tabular}[c]{@{}c@{}}Method\\ ----\\ Dataset\end{tabular}}} &
  \multicolumn{4}{c|}{\textbf{Proposed Heuristic}} &
  \multicolumn{4}{c|}{\textbf{Equalized Odds}} &
  \multicolumn{4}{c|}{\textbf{Calibrated Equalized Odds}} \\ \cline{2-13} 
 &
  \multicolumn{1}{c|}{\textbf{\begin{tabular}[c]{@{}c@{}}FPR\\ (Class 0)\end{tabular}}} &
  \multicolumn{1}{c|}{\textbf{\begin{tabular}[c]{@{}c@{}}FPR\\ (Class 1)\end{tabular}}} &
  \multicolumn{1}{c|}{\textbf{\begin{tabular}[c]{@{}c@{}}FNR\\ (Class 0)\end{tabular}}} &
  \multicolumn{1}{c|}{\textbf{\begin{tabular}[c]{@{}c@{}}FNR\\ (Class 1)\end{tabular}}} &
  \multicolumn{1}{c|}{\textbf{\begin{tabular}[c]{@{}c@{}}FPR\\ (Class 0)\end{tabular}}} &
  \multicolumn{1}{c|}{\textbf{\begin{tabular}[c]{@{}c@{}}FPR\\ (Class 1)\end{tabular}}} &
  \multicolumn{1}{c|}{\textbf{\begin{tabular}[c]{@{}c@{}}FNR\\ (Class 0)\end{tabular}}} &
  \multicolumn{1}{c|}{\textbf{\begin{tabular}[c]{@{}c@{}}FNR\\ (Class 1)\end{tabular}}} &
  \multicolumn{1}{c|}{\textbf{\begin{tabular}[c]{@{}c@{}}FPR\\ (Class 0)\end{tabular}}} &
  \multicolumn{1}{c|}{\textbf{\begin{tabular}[c]{@{}c@{}}FPR\\ (Class 1)\end{tabular}}} &
  \multicolumn{1}{c|}{\textbf{\begin{tabular}[c]{@{}c@{}}FNR\\ (Class 0)\end{tabular}}} &
  \multicolumn{1}{c|}{\textbf{\begin{tabular}[c]{@{}c@{}}FNR\\ (Class 1)\end{tabular}}} \\ \hline
\textbf{\begin{tabular}[c]{@{}c@{}} Online Fraud\\ (Country = 6)\end{tabular}} &
  \multicolumn{1}{c|}{0.001} &
  \multicolumn{1}{c|}{0.0009} &
  \multicolumn{1}{c|}{0.349} &
  \multicolumn{1}{c|}{0.246} &
  \multicolumn{1}{c|}{0.01} &
  \multicolumn{1}{c|}{0.009} &
  \multicolumn{1}{c|}{0.306} &
  \multicolumn{1}{c|}{0.282} &
  \multicolumn{1}{c|}{0.009} &
  \multicolumn{1}{c|}{0.01} &
  \multicolumn{1}{c|}{0.449} &
  \multicolumn{1}{c|}{0.124} \\ \hline
\textbf{\begin{tabular}[c]{@{}c@{}}Online Fraud\\ (Country = 22)\end{tabular}} &
  \multicolumn{1}{c|}{0.0007} &
  \multicolumn{1}{c|}{0.0008} &
  \multicolumn{1}{c|}{0.202} &
  \multicolumn{1}{c|}{0.13} &
  \multicolumn{1}{c|}{0.019} &
  \multicolumn{1}{c|}{0.021} &
  \multicolumn{1}{c|}{0.174} &
  \multicolumn{1}{c|}{0.113} &
  \multicolumn{1}{c|}{0.012} &
  \multicolumn{1}{c|}{0.021} &
  \multicolumn{1}{c|}{0.335} &
  \multicolumn{1}{c|}{0.104} \\ \hline
\textbf{\begin{tabular}[c]{@{}c@{}}Income\\ (AfricanAmerican)\end{tabular}} &  \multicolumn{1}{c|}{0.42} &
  \multicolumn{1}{c|}{0.351} &
  \multicolumn{1}{c|}{0.61} &
   \multicolumn{1}{c|}{0.77} &
  \multicolumn{1}{c|}{0.13} &
  \multicolumn{1}{c|}{0.12} &
  \multicolumn{1}{c|}{0.463} &
   \multicolumn{1}{c|}{0.348} &
  \multicolumn{1}{c|}{0.884} &
  \multicolumn{1}{c|}{0.345} &
  \multicolumn{1}{c|}{0.642} &
  \multicolumn{1}{c|}{0.705}
   \\ \hline  
\textbf{\begin{tabular}[c]{@{}c@{}}COMPAS\\ (AfricanAmerican)\end{tabular}} &
  \multicolumn{1}{c|}{0.387} &
  \multicolumn{1}{c|}{0.401} &
  \multicolumn{1}{c|}{0.521} &
  \multicolumn{1}{c|}{0.506} &
  \multicolumn{1}{c|}{0.435} &
  \multicolumn{1}{c|}{0.430} &
  \multicolumn{1}{c|}{0.492} &
  \multicolumn{1}{c|}{0.491} &
  \multicolumn{1}{c|}{0.315} &
  \multicolumn{1}{c|}{0.449} &
  \multicolumn{1}{c|}{0.561} &
  \multicolumn{1}{c|}{0.372} \\ \hline

\end{tabular}
}
\caption{Comparison to Equalized and Calibrated Equalized Odds.}
\vspace{-0.1in}
\label{tab:baselines}
\end{table*}

Next, we consider the exercise on the fraud detector model trained with XGBoost to sanitize across with two protected attributes country and currency. From Figure \ref{fig:sanxgboostfprbar}, we observe that we can achieve similar FPRs across the attribute values of currency. From a separate experiment (not presented for brevity), we also observe that sanitizing the model on currency inherently sanitizes it against country given their high dependence.

\textbf{Model benchmarks:} Next, we compare the proposed heuristic to Equalized  \cite{hardt2016equality} and Calibrated Equalized Odds \cite{pleiss2017fairness}. These techniques are aimed at achieving fairness across two groups (E.g. Country is/not 1), unlike the proposed heuristic which aims to achieve fairness across all sub groups with custom thresholds per country. In Table \ref{tab:baselines}, rows 1 and 2, we report FPR and FNR for Country is/not is Country 6\footnote{We pick the highest (Country 6) and lowest frequency (Country 22) country based on the transactions in our data set. Each country has at least 10,000 transactions each.} (Class 0 and Class 1). and Country is/not is Country 22 and observe that we are able to achieve similar metrics compared to the benchmarks. 


\textbf{Income-Prediction} This dataset from UCI Machine Learning Repository contains 14 demographic and occupational features for various people, with the goal of predicting whether a
person’s income is above \$50, 000. In this scenario, we seek to achieve predictions with equalized cost across genders (single protected attributes, two groups - Male and Female). In this dataset, we consider a scenario where the primary concern is ensuring equal generalized false negative rates across genders, which would help job recruiters prevent gender discrimination in salary estimates. Hence, we choose our fairness constraint to require relaxed equalized false negative rates across groups. In row 3, Table \ref{tab:baselines}, we observe that the proposed heuristic achieves similar FNRs for the two groups -- 0.77 (African American) and 0.61 (not African American). Also, note that we are able to achieve comparable (similar FNRs) or better performance (better FPRs) compared to Calibrated Equalized Odds, designed specifically to enforce equal FNRs across the two groups. Equalized Odds on the other hand, ensures equal FPRs and FNRs across the two groups, trading off the FPR for a lower FNR across groups (0.46 and 0.34) compared to the proposed heuristic (0.61 and 0.77). 

\textbf{Criminal Recidivism} Finally, we examine the proposed heuristic in t he context of criminal recidivism. As noted by several studies earlier, in this dataset, African Americans, receive
a disproportionate number of false positive predictions as compared with Caucasians when automated risk tools were used. Hence, we aim to equalize the generalized false positive rate. In row 4, Table \ref{tab:baselines}, we observe that the proposed heuristic achieve a) similar FPRs for the two groups -- 0.4 (African American) and 0.38 (not African American) b) similar FNRs (0.60 and 0.59 respectively). Note that we are able to achieve comparable (similar FNRs) and slightly better better FPRs compared to Equalized Odds, designed to enforce equal FNRs and FPRs across the two groups. When compared to the Calibrated Equalized Odds, where the weighted combination of error rates are matched (equal FPRs and FNRs as discussed in \cite{pleiss2017fairness}), we observe that the proposed heuristic performs comparably in terms of both FPRs and FNRs across two groups.


\vspace{-0.1in}
\section{Conclusion}

We propose a fairness measure relaxing the FPR and TPR equality conditions in the popular equal odds fairness regime \cite{hardt2016equality}. To conform to the proposed fairness, we design an iterative, model-agnostic, grid-based heuristic that calibrates the outcomes per sensitive attribute value (for e.g., different countries). Through a detailed case study of our motivating application, fraud detection, we show that the proposed heuristic is able to achieve fairness across multiple values of a single protected attribute, multiple protected attributes. We  compare our work to current fairness techniques and show comparable performance across several public data sets.

\vspace{-0.1in}
\bibliographystyle{ACM-Reference-Format}
\bibliography{sample-base}


\end{document}